\documentclass{article} 
\usepackage[preprint]{colm2026_conference}

\usepackage{microtype}
\usepackage{hyperref}
\usepackage{url}
\usepackage{booktabs}
\usepackage{amsmath}
\usepackage{array}

\usepackage{lineno}

\definecolor{darkblue}{rgb}{0, 0, 0.5}
\hypersetup{colorlinks=true, citecolor=darkblue, linkcolor=darkblue, urlcolor=darkblue}

\newcommand{\system}{\textsc{PlaceMem}}

\title{\system: Toward a Compute-Aware Memory Plane for Lifelong Agents}

\author{%
  Sukanta Ganguly, Ph.D \\
  \texttt{sukantag@netapp.com}
}

\begin{document}

\raggedbottom

\maketitle

\begin{abstract}
Lifelong agents need more than larger context windows and better retrieval. They need memories that can persist, evolve, and be corrected without forcing the serving stack to recompute the same history on every turn or silently reuse stale runtime state. We present \system{} as a systems position on lifelong-agent memory, instantiated by an executable control-plane prototype. The central claim is that agent memory should be represented as versioned \emph{capsules} that unify semantics, provenance, validity, and reusable runtime state under one correction-aware identity. In the current prototype, capsules drive prompt-level text retrieval, KV-aware routing, and cascading invalidation over live streamed backends; prospective layer-frontier replay is intentionally framed as a deeper integration agenda rather than a claimed engine feature. We describe a vLLM-first prototype with persistent capsule state, concurrency-safe invalidation, an OpenAI-compatible routing sidecar, a typed metadata contract, and a benchmark harness that measures live first-token latency, reuse, and post-correction behavior. The result is both an executable artifact that demonstrates correction-aware control-plane behavior today and a concrete roadmap for replay-aware serving integration in future lifelong-agent systems.
\end{abstract}

\section{Introduction}

Lifelong agents are emerging as a unifying target for language-model systems: they must remember users, refine beliefs, revise plans, accumulate tool experience, and stay responsive under repeated interaction. Recent systems have shown that memory improves planning, reflection, and multi-session behavior for language agents \citep{park2023generative, shinn2023reflexion, packer2024memgpt}. In parallel, serving systems such as vLLM have substantially improved the runtime efficiency of large language model (LLM) inference through better KV management and batching \citep{kwon2023vllm}. However, these two lines of work still optimize disjoint layers of the stack, which makes it difficult to evaluate memory quality and systems efficiency within one coherent architecture.

The first layer is \emph{semantic memory}: facts, summaries, tool traces, episodic reflections, and user-specific context retrieved for the next turn. The second is \emph{runtime memory}: prefix caches, KV tensors, offloaded blocks, and placement hints used by the serving system. Lifelong-agent research usually treats the second layer as an opaque backend. Serving research usually treats the first as unstructured prompt text. As a result, repeated multi-turn agent workloads suffer from two avoidable failures: they recompute expensive prefills even when semantically relevant reusable state exists, and they continue to expose stale compute artifacts after a fact, summary, or plan has been corrected. We believe this memory--inference boundary is becoming a first-order bottleneck for the next generation of agent systems.

This paper introduces \system{} as a systems position on lifelong-agent memory, backed by an executable control-plane prototype. Its core abstraction is a \emph{memory capsule}, a versioned object that binds semantic content and reusable compute artifacts under a single identity. Capsules drive replay selection, reuse eligibility, and cascading invalidation across distributed inference nodes. More broadly, \system{} is intended as a proposal for how lifelong-agent memory should be operationalized when inference itself is distributed, latency-sensitive, and continuously reused. The current paper does not claim a finished replay engine; instead, it demonstrates correction-aware control-plane behavior today and uses deeper replay integration as the agenda that follows from that result.

\paragraph{Contributions.}
We make three contributions.
\begin{itemize}
   \item We argue for a compute-aware memory plane as a missing systems abstraction for lifelong agents, and define memory capsules as its correction-aware core unit.
   \item We present an executable vLLM-first prototype that already demonstrates correction-aware control-plane behavior through persistent capsule state, concurrency-safe invalidation, an OpenAI-compatible sidecar, a typed serving contract, and a live benchmark harness.
   \item We define a replay-and-placement policy, together with an explicit systems agenda for deeper integration, that trades off semantic value, latency, staleness risk, and locality across text retrieval and KV replay while leaving layer replay as future engine work rather than a claimed implementation result.
\end{itemize}

\section{Why lifelong-agent memory needs a systems layer}

The central systems problem is not only \emph{what} an agent should remember, but also \emph{where} that memory should live and \emph{how} it should be invalidated once the world changes. Consider a customer-support agent that first stores an outdated refund policy, then later receives a corrected policy. A text-only memory system can replace the sentence that will be retrieved next time, but it usually does not know about derivative summaries, graph edges, prefix caches, or remote KV artifacts that were created from the stale policy. Conversely, a serving stack may reuse runtime state efficiently, but lacks a principled view of whether the underlying semantics remain valid.

Lifelong agents therefore need a missing middle layer with four properties:
\begin{itemize}
   \item \textbf{Unified identity.} The same object should name both memory semantics and compute artifacts derived from them.
   \item \textbf{Freshness-aware reuse.} Reuse should be blocked when validity windows, provenance, or contradiction signals indicate elevated risk.
   \item \textbf{Locality-aware execution.} Requests should move toward reusable artifacts when that is better than fetching them remotely or recomputing them.
   \item \textbf{Cascading forgetting.} Corrections and deletions should invalidate derived summaries, indexes, and runtime artifacts rather than merely updating a text row.
\end{itemize}

\begin{table}[t]
\centering
\small
\begin{tabular}{lcccc}
\toprule
Approach & Semantics & Reuse & \shortstack{Cascade\\invalidate} & \shortstack{Placement\\aware} \\
\midrule
Prompt or vector memory & Yes & No & Rarely & No \\
Runtime prefix or KV cache & No & Yes & No & Local only \\
\system{} capsules & Yes & Yes & Yes & Yes \\
\bottomrule
\end{tabular}
\caption{A qualitative comparison of text-centric memory, runtime caches, and \system.}
\label{tab:comparison}
\end{table}

Table~\ref{tab:comparison} summarizes the distinction. \system{} does not replace retrieval or runtime caching. Instead, it makes them interoperable and auditable for lifelong agents.

\section{The \system{} architecture}

\begin{figure}[t]
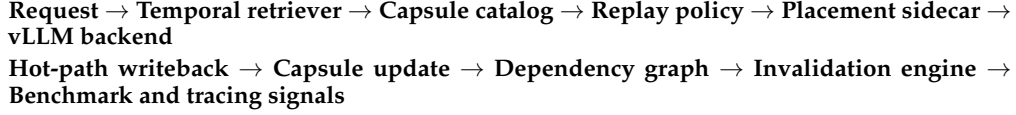

\centering
\setlength{\fboxsep}{8pt}
\fbox{\parbox{0.95\linewidth}{
\small
\textbf{Request} $\rightarrow$ \textbf{Temporal retriever} $\rightarrow$ \textbf{Capsule catalog} $\rightarrow$ \textbf{Replay policy} $\rightarrow$ \textbf{Placement sidecar} $\rightarrow$ \textbf{vLLM backend} \\
[2pt]
\textbf{Hot-path writeback} $\rightarrow$ \textbf{Capsule update} $\rightarrow$ \textbf{Dependency graph} $\rightarrow$ \textbf{Invalidation engine} $\rightarrow$ \textbf{Benchmark and tracing signals}
}}
\caption{\system{} inserts a control plane between agent memory and distributed inference.}
\label{fig:arch}
\end{figure}

Figure~\ref{fig:arch} shows the control path. The architecture is organized around a capsule catalog, a dependency graph, and a scheduler that reasons jointly about semantic relevance and runtime reuse.

\subsection{Memory capsules}

A memory capsule is a versioned record with required semantic fields such as tenant identity, surface text, timestamps, and validity windows, together with optional enrichment such as summaries, provenance, entities, facts, dependencies, and artifact locators. The crucial step is that a capsule can also name reusable runtime artifacts such as embeddings, KV cache segments, and optional layer-frontier checkpoints.

This design gives one identifier for meaning and compute. A retriever can therefore answer not only \emph{which memory is relevant}, but also \emph{which reusable state is compatible and cheap to consume on a specific node}. In the current prototype, text and KV-aware routing are fully wired end to end, while layer-frontier replay remains an explicit contract for future engine extensions.

\subsection{Replay policy and placement}

For each request, \system{} ranks replay options across prompt-level text retrieval, KV replay, and prospective layer replay. The policy evaluates a capsule $c$ on target $t$ under replay mode $r$ with a utility function of the form
\begin{equation}
U(c, t, r) = w_1 P_{\text{reuse}} + w_2 C_{\text{saved}} + w_3 V_{\text{sem}} - w_4 R_{\text{stale}} - w_5 B_{\text{remote}} - w_6 L_{\text{target}}.
\end{equation}

Here, $P_{\text{reuse}}$ estimates near-term reuse, $C_{\text{saved}}$ measures avoided prefill work, $V_{\text{sem}}$ captures task relevance, $R_{\text{stale}}$ captures freshness risk, $B_{\text{remote}}$ estimates transfer cost, and $L_{\text{target}}$ penalizes latency-sensitive targets. The selected decision is then turned into backend metadata for the serving layer.

This differs from a local prefix cache in two ways. First, the decision is made in a control plane rather than inside a single worker. Second, the decision is conditioned on semantic validity and locality, not only on byte reuse. In other words, \system{} treats memory reuse as a policy question, not merely a storage optimization. In the current prototype, the scheduler exercises this policy over a configurable node topology; multi-node heterogeneous placement evaluation is a forward agenda item.

\subsection{Invalidation as a correctness primitive}

For lifelong agents, forgetting is not a convenience feature; it is part of staying behaviorally current after corrections. When a fact is corrected, or a plan is superseded, the agent should not continue to reuse runtime state that was derived from the stale information. \system{} therefore maintains dependency edges between capsules and derived artifacts, allowing an invalidation event to cascade through summaries, indexes, graph facts, and reusable compute state.

This mechanism turns memory correction into a first-class operational event. In practice, it creates an auditable answer to a question that many agent stacks leave implicit: after a correction, which downstream artifacts are no longer safe to reuse?

\section{A vLLM-first prototype}

We implement \system{} as a concrete control-plane prototype anchored on vLLM, intended as executable evidence for the systems position above rather than a claim that all replay paths are native to the serving engine. The implementation includes: a capsule schema, a catalog, a dependency graph, persistent state snapshots, concurrency-safe invalidation, a placement RPC contract, an HTTP control plane, a vLLM-facing admission shim, a multi-node OpenAI-compatible sidecar, and a benchmark harness with repeated-run JSON and CSV exports.

The sidecar provides an immediately deployable integration surface with existing vLLM-compatible endpoints and establishes a stable typed-header contract for deeper engine integration, keeping the control-plane behavior deployable today while exposing the exact seam where replay-aware integration can be added later.

The present implementation is deliberately control-plane heavy. It directly supports text and KV-aware routing over live streamed backends, while exposing layer-checkpoint artifacts and replay mode selection as a forward-compatible contract. This gives a practical evaluation surface for correction-aware reuse and invalidation without requiring an invasive engine fork.

\section{Pilot evaluation on lifelong-agent workloads}

We evaluate on three workload families that stress distinct failure modes of lifelong memory: customer-support history, coding-agent repository loops, and research-agent fact revision. The goal of this section is intentionally narrow: validate correction-aware control-plane behavior on repeated lifelong-agent traces, not claim that replay-aware serving integration is already complete.

We compare four baselines on the same 48-turn pilot trace: prompt-only execution, text-only retrieval that prepends a retrieved capsule summary without exposing reusable KV artifacts, runtime reuse without semantic invalidation, and full \system.

The text-only baseline is intentionally prompt-layer, because that is how current agent-memory stacks typically consume retrieved text when no reusable compute artifact is available. The pilot preserves the real control-plane, admission, and sidecar path, but replaces the final token generator with a replay-mode-sensitive streaming mock (a fixed SSE payload with a configurable inter-token delay, so TTFT reflects only control-plane capsule lookup and sidecar routing overhead) to isolate cross-layer reuse and invalidation behavior before full cluster-scale vLLM evaluation. We therefore use this pilot to test the correction-aware control-plane thesis directly, while treating deeper replay-aware engine integration as subsequent work.

In the current pilot, we report mean time-to-first-token, replay reuse rate, post-correction stale-hit counts, and invalidation overhead, because those are fully executable end to end today and directly test whether correction-aware control-plane behavior changes the correctness--latency trade-off of reuse.

\begin{table}[ht]
\centering
\small
\begin{tabular}{lrrrr}
\toprule
Baseline & Mean TTFT & $\Delta$ TTFT & Reuse & Stale hits \\
\midrule
Prompt-only & 18.25 & --- & 0.00 & 0 \\
Text-only & 18.31 & +0.3\% & 0.00 & 0 \\
Runtime-only & 6.72 & -63.2\% & 1.00 & 17 \\
Full \system{} & 7.17 & -60.7\% & 1.00 & 0 \\
\bottomrule
\end{tabular}
\caption{Aggregate pilot comparison over 48 turns spanning support, coding, and fact-revision traces using the live control plane, routing sidecar, and a replay-mode-sensitive mock backend.}
\label{tab:pilot-results}
\end{table}

Table~\ref{tab:pilot-results} sharpens the main systems claim. Relative to prompt-only execution, both runtime-only reuse and full \system{} reduce mean time-to-first-token by about 61--63\% on this controlled backend. The crucial distinction is correction handling: disabling invalidation preserves the latency win but yields 17 stale post-correction reuses, whereas full \system{} retains the latency benefit while eliminating stale hits entirely. Mean invalidation overhead in the full system is 1.09 ms over three correction events. \emph{Note: TTFT here measures control-plane and sidecar routing latency against the streaming mock; values do not reflect real prefill savings on a production GPU.}

\begin{table}[ht]
\centering
\scriptsize
\begin{tabular}{lrrrr}
\toprule
Condition & Mean TTFT (ms) & KV reuse & Post-corr.\ acc.\ & Stale sel.\ \\
\midrule
Text-summary baseline & 36.08 & 0.00 & 1.00 & --- \\
Reuse w/o invalidation & 20.73 & 1.00 & 0.00 & 1.00 \\
Full \system{} & 20.54 & 1.00 & 1.00 & 0.00 \\
\midrule
\multicolumn{5}{l}{\itshape Deployment throughput (340-turn batch, 3 repeats)} \\
\itshape Full \system{} & \itshape 22.42 & \itshape 1.00 & \itshape --- & \itshape --- \\
\bottomrule
\end{tabular}
\caption{Real-backend comparison on LiquidAI/LFM2.5-230M. Top three rows: live correction-follow probe (9 post-correction queries per condition $\times$ 3 workload families = 27 post-correction turns per condition). Bottom row: deployment-throughput validation across 3 repeats of the full 340-turn benchmark. Post-correction accuracy measures whether the live model returns the corrected answer; stale selection measures whether the control plane selects an invalidated capsule.}
\label{tab:deployment-validation}
\end{table}

Table~\ref{tab:deployment-validation} integrates both the latency and behavioral stories on the real vLLM backend. The correction-follow probe uses a controlled task: each scenario has a structured canonical identifier injected into the active capsule summary, and accuracy measures whether the live model's response contains the post-correction identifier rather than the stale one, isolating whether the control plane forwards the right summary rather than testing open-ended generation quality. The three-condition comparison shows why invalidation matters: without it the control plane selects the stale capsule after a correction event, the model receives the outdated summary, and post-correction accuracy drops to zero. Full \system{} eliminates stale selection while matching the TTFT of the reuse-without-invalidation condition (20.54 ms versus 20.73 ms), confirming that correction-aware control adds faithfulness at no latency cost. The text-summary baseline reaches full accuracy but at 36.08 ms mean TTFT, a 75\% overhead relative to full \system{}. The deployment-throughput row confirms the prototype runs end to end on a real vLLM deployment.

\section{Related work}

Agent-memory systems (Generative Agents, Reflexion, MemGPT) model what to remember and retrieve, but leave runtime reuse and invalidation to the serving layer \citep{park2023generative, shinn2023reflexion, packer2024memgpt}. Serving systems such as vLLM optimize KV memory, batching, and placement, but do not make semantic freshness or correction first-class control-plane inputs \citep{kwon2023vllm}.

Semantic- and context-caching systems are the closest neighbors \citep{bang2023gptcache, li2024scalm, gao2025onlinecontext}. GPTCache and SCALM match on similarity but carry no validity windows or dependency edges; Online Context Caching optimizes prefix-block locality but treats blocks as opaque artifacts with no correction semantics. Neither exposes a versioned object jointly carrying meaning, validity, reusable artifacts, and edges for correction-driven invalidation.

\system{}'s cross-layer claim is that a single capsule identity coordinates retrieval, reuse eligibility, and correction-driven invalidation.

\paragraph{Conclusion.}

Lifelong agents will not be reliable if semantic memory and runtime reuse remain weakly synchronized. \system{} closes this gap with a versioned capsule design backed by a working prototype, giving future researchers a direct surface for validating KV replay gains, multi-node placement, and correction-aware invalidation at scale.

\paragraph{Ethics Statement.}

Persistent agent memory raises privacy and misuse concerns. We treat provenance, tenant scoping, and explicit invalidation as mandatory metadata to make retention and correction observable.

\bibliography{colm2026_conference}
\bibliographystyle{colm2026_conference}

\end{document}